\documentclass[11pt,a4paper]{article}
\usepackage[hyperref]{emnlp2020}
\usepackage{times}
\usepackage{latexsym}

\usepackage{fullpage}
\usepackage[top=2cm, bottom=4.5cm, left=2.5cm, right=2.5cm]{geometry}
\usepackage{amsmath,amsthm,amsfonts,amssymb,amscd}
\usepackage{lastpage}
\usepackage{enumerate}
\usepackage{fancyhdr}
\usepackage{mathrsfs}
\usepackage{xcolor}
\usepackage{graphicx}
\usepackage{adjustbox}
\usepackage{caption}
\usepackage{subcaption}
\usepackage{listings}
\usepackage{hyperref}
\usepackage{enumitem}
\usepackage{float}
\usepackage{fancyvrb}
\usepackage{enumitem}
\usepackage{bbold}
\usepackage[T1]{fontenc}
\usepackage{microtype}
\usepackage{graphicx}
\aclfinalcopy % Uncomment this line for the final submission
\title{EfficientQA: a RoBERTa Based Phrase-Indexed Question-Answering System}

\author{Sofian Chaybouti, Achraf Saghe, Aymen Shabou \\
\textit{DataLab Groupe, Crédit Agricole S.A} \\
Montrouge, France\\}

\begin{document}
\maketitle

\begin{abstract}

State-of-the-art extractive question-answering models achieve superhuman performances on the SQuAD benchmark. Yet, they are unreasonably heavy and need expensive GPU computing to answer questions in a reasonable time. Thus, they cannot be used in the open-domain question-answering paradigm for real-world queries on hundreds of thousands of documents. In this paper, we explore the possibility of transferring the natural language understanding of language models into dense vectors representing questions and answer candidates to make the question-answering task compatible with a simple nearest neighbor search task. This new model, which we call \textit{EfficientQA}, takes advantage of the pair of sequences kind of input of BERT-based models \cite{devlin2019bert} to build meaningful dense representations of candidate answers. These latter are extracted from the context in a question-agnostic fashion. Our model achieves state-of-the-art results in \textit{Phrase-Indexed Question Answering} (PIQA) \cite{seo2018phraseindexed} beating the previous state-of-art \cite{seo-etal-2019-real} by 1.3 points in exact-match and 1.4 points in f1-score. These results show that dense vectors can embed rich semantic representations of sequences, although these were built from language models not originally trained for the use case.
Thus, to build more resource-efficient NLP systems in the future, training language models better adapted to build dense representations of phrases is one of the possibilities.
\end{abstract}

\section{Introduction}

\begin{figure}[H]
\centering
    \includegraphics[width = 0.45\textwidth]{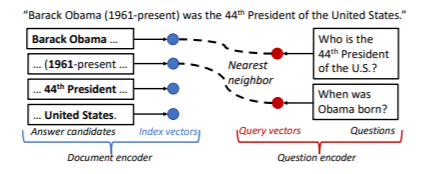}
\caption{The PIQA challenge from \cite{seo2018phraseindexed}}
\label{piqa}
\end{figure}

Question answering is the discipline that aims to build systems that automatically answer questions posed by humans in a natural language. In the extractive question-answering paradigm, the answers are spans of text extracted from a single document. In the famous SQuAD benchmark \cite{rajpurkar2016squad}, for instance, each answer lies in a paragraph from Wikipedia. \\
In the open-domain setting, the answers are sought in a large collection of texts such as the whole English Wikipedia \cite{chen2017reading}. State-of-the-art performances in usual Question Answering are achieved thanks to powerful and heavily pre-trained language models that rely on sophisticated attention mechanisms and hundreds of millions of parameters. Attention mechanisms \cite{bahdanau2016neural} are key components of such systems since they allow building contextualized and question-aware representations of the words in the documents and extracting the span of text, which is most likely the correct answer. These models are very resource-demanding and need GPUs to be scalable. Thus, they seem unsuitable to the open-domain real use cases, where the model has to be applied on hundreds of thousands of documents, even with a multi-GPU server.\\
A first approach to solve this issue would be first applying a filter based on a statistical algorithm like \textit{tf-idf} \cite{10.5555/106765.106782} vectors or \textit{BM25} \cite{https://doi.org/10.1002/asi.4630270302} algorithm. Then, the heavy model is called on several dozens of paragraphs. This approach is still prohibitive with CPU-only resources, for instance.  \\
\cite{seo2018phraseindexed} introduces a new benchmark, called \textit{Phrase-indexed Question Answering}, which constrains the usual extractive question-answering task. Indeed, document and question encoders are forced to be independent (figure \ref{piqa}). First, a document is processed to provide a vector representation to each answer candidate in an offline mode. Then, the query is processed in the online step to be mapped to its vector representation. Hence, the answer to the query is obtained by retrieving the nearest candidate vector to the query vector. The general form of such an approach to solve the open-domain question could be reformulated as follows. First, all candidate answers from all documents in the database are indexed offline. Then, the question is encoded at inference time, and a simple nearest-neighbor search retrieves the best candidate. This way, the scalability challenge of QA systems is improved since a single pass forward in the deep learning model is needed to encode the question instead of several ones (one per document) in previous settings. \\
This paper proposes a new algorithm to solve the PIQA benchmark (figure \ref{piqa}) and close the gap between classic QA models. Our approach takes advantage of BERT-based models in two ways. First, it extracts the potential answer candidates in a question-agnostic fashion. Secondly, it takes two sequences as input to build powerful semantic representations of candidate answers. Finally, it trains a Siamese network to map candidates' answers and queries in the same vector space. \\
Our model performs well, beating DENSPI \cite{seo-etal-2019-real}, the previous state-of-the-art on the PIQA benchmark, by 1.4 points in f1-score and 1.3 points in exact-match, while being less resource-demanding both in training and at inference times. It requires indexing only a hundred answer-candidates dense vectors per context and finetuning a RoBERTa-based \cite{liu2019roberta} model, while DENSPI uses a BERT-large model.  

\section{Background}

\subsection{Machine Reading Comprehension}

The construction of vast Question Answering datasets, particularly the SQuAD benchmark \cite{rajpurkar2016squad}, has led to end-to-end deep learning models successfully solving this task, for instance, \cite{seo2018bidirectional} is one of the first end-to-end model achieving impressive performances. More recently, the finetuning of powerful language models like \textit{BERT} \cite{devlin2019bert} allowed achieving superhuman performances on this benchmark.
In \textit{SpanBERT} \cite{joshi2020spanbert}, the pretraining task of the language model is masked span prediction instead of masked word prediction to be better adapted to the downstream task of QA, which consists of span extraction.
All these models rely on the same paradigm: building query-aware vector representations of the words in the context. This fundamental idea makes these models unsuitable for the Open-Domain setting. 

\subsection{Open-Domain Question Answering}

\cite{chen2017reading} introduced the Open-Domain Question Answering setting that aims to use the entire English Wikipedia as a knowledge source to answer factoid natural language questions. This setting brings the challenge of building systems able to perform \textit{Machine Reading Comprehension} at scale. \\
Most recent work explored the following pipeline to solve this task. First, dataset documents are indexed (or encoded) using statistical methods like \textit{BM25} or dense representations of documents. Then, we retrieve dozens of them by similarity search between documents and questions \cite{karpukhin2020dense}. Finally,  we apply a deep learning model trained for machine reading comprehension to find the answer. This approach has been developed in a number of papers \cite{chen2017reading}, \cite{raison2018weaver}, \cite{min2018efficient},  \cite{wang2017r3}, \cite{lee2018ranking}, \cite{Yang_2019}. It takes advantage of SOTA's very powerful language models but has the inconvenience of being resource-demanding. Moreover, its performances are capped by the capabilities of the document retrieval step of the pipeline.

\subsection{the PIQA Challenge}

 \cite{seo2018phraseindexed} introduced the \textit{Phrase-Indexed Question Answering} (PIQA) benchmark to make machine reading comprehension scalable. This benchmark enforces independent encoding of question and document answer candidates to reduce the question-answering task to a simple similarity search task. Closing the gap between such systems and powerful models relying on query-aware context representation would be a great step toward solving the open-domain question-answering scalability challenge. The baselines proposed use LSTM encoders trained in an end-to-end fashion. While achieving encouraging results, the performances are far from state-of-the-art attention-based models.
 
\textit{DENSPI} \cite{seo-etal-2019-real} is the current state-of-the-art on the PIQA benchmark. This system uses the BERT-large language model to train a siamese network able to encode questions and indexed answer candidates independently. To represent candidate answers, \textit{DENSPI} builds dense representations using each index phrase's start and end positions. \textit{DENSPI} is also evaluated on the SQuAD-open benchmark \cite{chen2017reading}. While being significantly faster than other systems, it needs to be augmented by sparse representations of documents to be on par with them in terms of performance. 

\textit{Ocean-Q} \cite{fang2020accelerating} proposes an interesting approach to solve both the PIQA and the Open-Domain QA benchmarks by building an ocean of question-answer pairs using Question Generation and query-aware QA models. When a question is asked, the most similar question from the ocean is retrieved thanks to token similarity. This approach avoids the question-encoding step while being on par with previous models on the SQuAD-open benchmark and significantly higher than the baselines on the PIQA challenge.  

\section{Model}
In this section, the model and the algorithm to solve the task are developed. 

\subsection{Problem Definition}

The problem tackled in this paper is \textit{Phrase-indexed Question Answering}. Vanilla Question Answering is the task of building systems that can answer natural language questions with spans of text in the documents (figure with example \ref{piqa}). Formally, the goal is to design a function $F$ mapping a question $Q$  and a context $C$, both represented by a sequence of tokens $\{q_1, q_2, ..., q_n\}$ and $\{c_1, c_2, ..., c_m\}$ respectively, to a subsequence of $C$ as an answer $A = \{a_1, a_2, ..., a_p\}$ (eq. \ref{eq:qa}).
\begin{equation}
F(Q, C) = A
\label{eq:qa}
\end{equation}
In \textit{PIQA}, $F$ is constrained to be an $argmax$ over a set of answer candidates $\{A_1, A_2, ..., A_k\}$ ($k$ subsequences of the context $C$) of a similarity product between the encoding $G(Q) \in \mathbf{R}^l$ of the question and the encoding of each candidate $H(A_i) \in \mathbf{R}^l$ (eq. \ref{eq:argmax}), where $l$ is the encoding size.
\begin{equation}
A = {argmax}_{A_i} G(Q) \bullet H(A_i)
\label{eq:argmax}
\end{equation}

\subsection{Agnostic Extraction of Answer Candidates}
The first step toward building the system is defining the answer candidates. A naive approach would be to consider all possible spans in a given context $C$ of length $m$. This would give $\frac{m (m + 1)}{2}$ possible candidates, i.e., about $10^5$ candidates per context if we assume contexts are made of about 500 tokens.

\begin{figure}[H]
    \includegraphics[width = 0.48\textwidth]{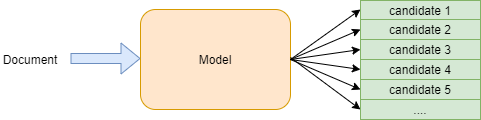}
\caption{Agnostic extraction of answer candidates}
\label{agnostic_base}
\end{figure}
Only a limited amount of all possible spans are potential answers to any question. Thus, we reduce the set of candidates by training a Question-Agnostic Answer Candidates Extraction model (figure \ref{agnostic_base}). Formally, the context $C$ is mapped to the set of candidates $\{A_1, ..., A_k\}$ thanks to a function $f$.

\begin{figure*}[ht]
    \includegraphics[width = \textwidth]{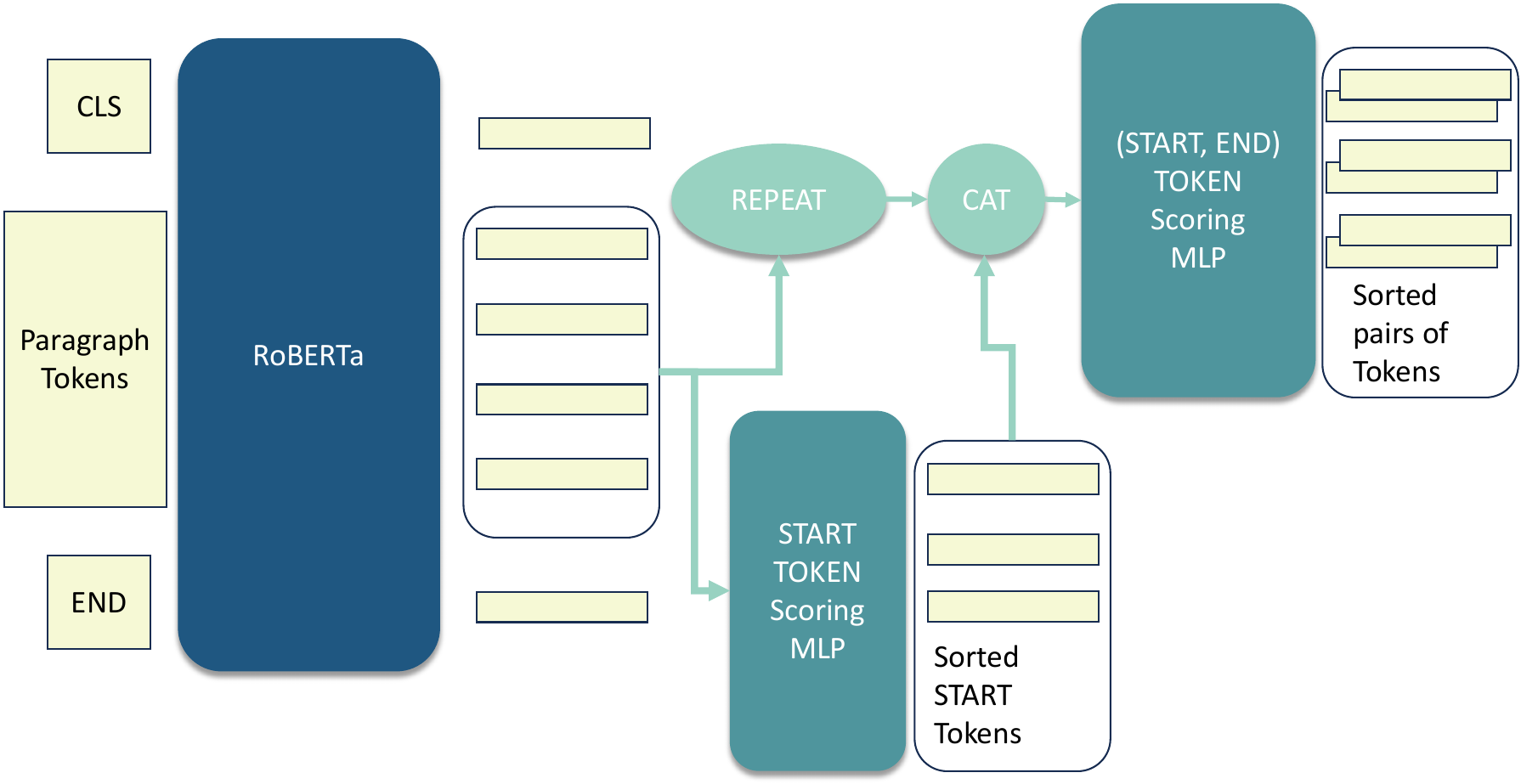}
\caption{Agnostic extraction of answer candidates with beam search. Paragraph tokens are provided to the language models to produce their embeddings, then a first dense layer allows to identify the $s$ most likely start positions of candidates. The embeddings of the paragraph's tokens are concatenated to each start position, and a second dense layer identifies the $e$ most likely end positions associated with each start position. We end up with $s \times e$ possible spans.}
\label{agnostic}
\end{figure*}

To do so, a \textit{Roberta Base} \cite{liu2019roberta} model is finetuned, taking the context as input and supervised by the answers provided in the SQuAD dataset. We use a beam search algorithm to extract the candidates: the $s$ most likely candidate starts are first extracted thanks to a dense layer. Then, their embeddings are concatenated to each context word embeddings and fed into another dense layer to extract the $e$ most likely candidate ends associated to each start position as shown in figure \ref{agnostic}. Thus, we end up processing $s \times e$ answer candidates. Ablation studies, developed in further sections, show that feeding the start position embeddings when extracting the end positions results gives better answer candidates.

\subsection{Building of dense vectors of answer candidates and questions}

After defining the set over which the $argmax$ function will be applied, we need to build the encoding functions for both the questions and the candidate's answers. To this purpose, we finetune a \textit{Roberta Base} as a \textit{siamese network} \cite{Koch2015SiameseNN} so that questions and candidates are mapped to the same euclidean space (eq. \ref{eq:dense}).
\begin{equation}
G \simeq H
\label{eq:dense}
\end{equation}

\subsubsection{Answer Candidates Dense Representations}

\begin{figure}[ht]
    \centering
    \includegraphics[width = 0.4\textwidth]{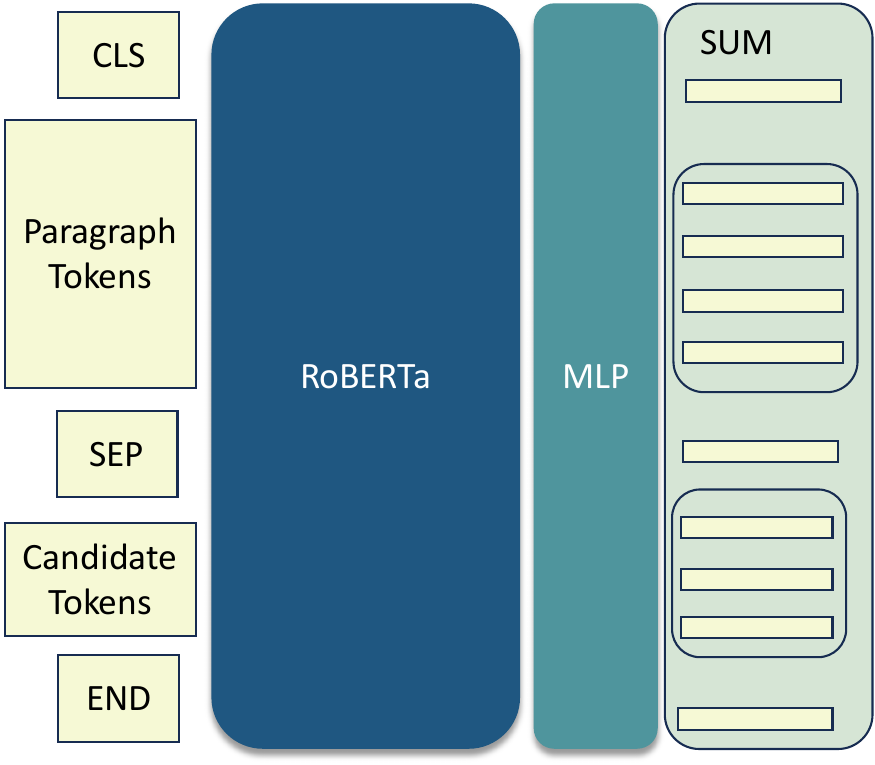}
\caption{Answer candidate dense vectors. The paragraph's tokens and the candidate's tokens separated by special \textit{SEP} token are provided to the language model. The final embeddings are provided to an additional dense layer and averaged to produce the candidate dense representation.}
\label{adv}
\end{figure}

We take advantage of the pair of sequences type of input of pre-trained BERT-based models to build powerful answer candidate representations. The context is provided as the first input and the candidate as the second input of the encoder, as shown in figure \ref{adv}. Eventually, the embeddings of each token are passed through a dense layer, and their final embeddings are averaged to provide the encoding of the candidate.

\subsubsection{Question Dense Representation}

\begin{figure}[ht]
   \centering
    \includegraphics[width = 0.4\textwidth]{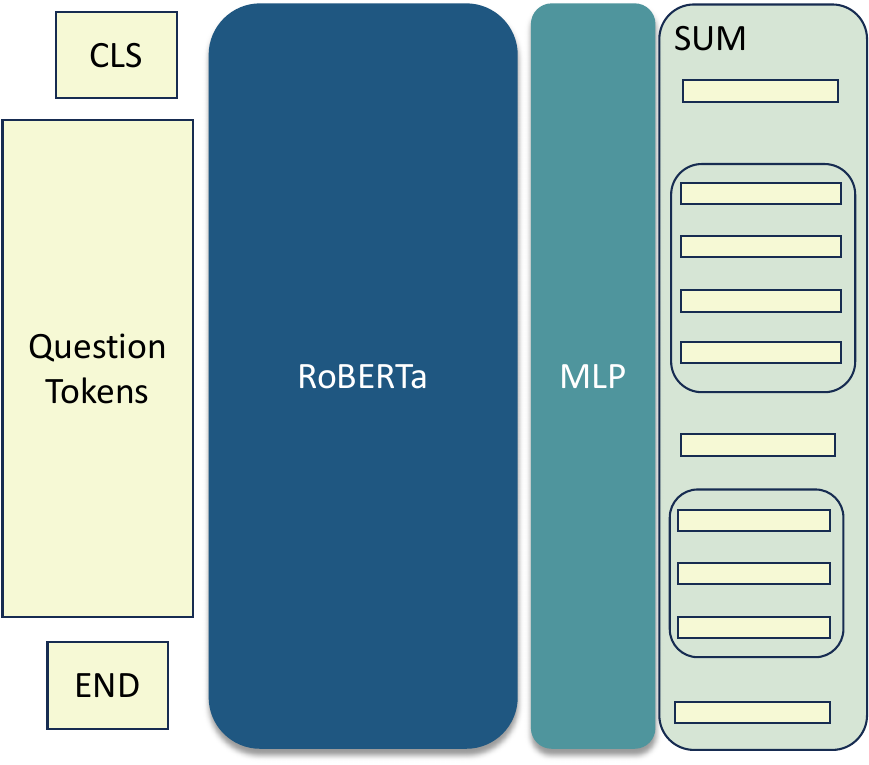}
\caption{Question dense vectors. The question's tokens are passed to the same language model as the candidates', and the final embeddings are passed to the same dense layer. Eventually, the vectors are averaged to produce the question-dense representation.}
\label{qdv}
\end{figure}

The question is passed through the same network as the context-candidate pair to build its representation, and the embeddings of all tokens are averaged as shown in figure \ref{qdv}. 

\subsection{Training Objectives}
\subsubsection{Candidates Extraction}

When training the Question-Agnostic Candidates Extraction model, we use the cross-entropy loss over start and end positions, just like most of deep neural networks trained for vanilla Question Answering but without adding the question information as described in eq.\ref{eq:agnostic}.

\begin{equation}
\begin{adjustbox}{max width=0.5\textwidth}
$
L(C; \mathbf{\Theta}) = - log(\mathbf{P}(s^*; \mathbf{\Theta})) - log(\mathbf{P}(e^*; \mathbf{\Theta}))
$
\label{eq:agnostic}
\end{adjustbox}
\end{equation}

\subsubsection{Phrase-Indexed Question-Answering}
We use the candidates extracted previously to train the Siamese network to build the questions' and candidates' representations. When the correct answer $A^*$ is among these candidates, the loss described in eq.\ref{eq:piqa}, where $\Gamma$ represents the parameters of the networks, is minimized.

\begin{multline}
L(Q, A_i; \Gamma) = - H(A^*) \bullet G(Q)\\ 
+ log(\sum_i \textit{exp}(H(A_i) \bullet G(Q))
\label{eq:piqa}
\end{multline}

\section{Experiment}
In this section, we present our experiments and results. 

\subsection{Data}

\subsubsection{SQuAD v1.1}

\begin{figure}[ht]
    \includegraphics[width = 0.48\textwidth]{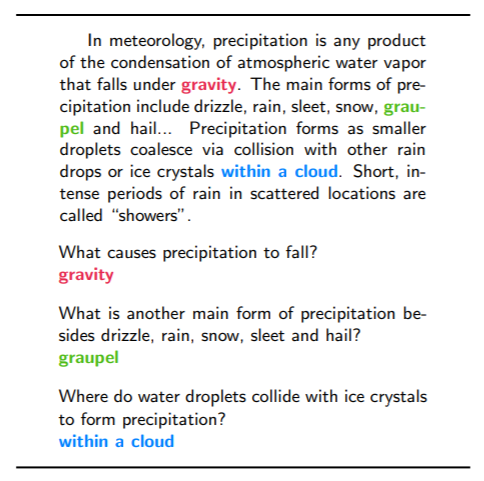}
\caption{Question-answer pairs for a passage in the
SQuAD dataset (figure taken from \cite{rajpurkar2016squad})}
\label{squad}
\end{figure}

SQuAD v1.1 (figure \ref{squad}) \cite{rajpurkar2016squad} is a reading comprehension dataset consisting of 100,000+ question-answers pairs from Wikipedia paragraphs. Our model was trained on the train set ($87599$ pairs) and evaluated on the development set ($10570$ pairs).

\subsubsection{FQuAD: The French Question Answering Dataset}

\begin{figure}[ht]
    \includegraphics[width = 0.48\textwidth]{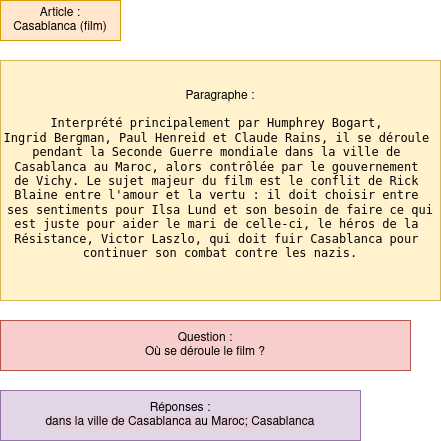}
\caption{Question-answers pair for a given paragraph in the
FQuAD dataset.}
\label{fquad}
\end{figure}

In recent years, efforts have been made to democratize NLP powerful tools beyond the English language. To this purpose, new datasets in other languages have been designed. The French Question Answering Dataset, called FQuAD (figure \ref{fquad})) is one of them \cite{dhoffschmidt2020fquad}. FQuAD is a French question-answering corpus built from 326 Wikipedia articles, with train and development sets consisting of 20,731 and 5,668 question/answer pairs, respectively.

\subsection{Training Details}

\subsubsection{Agnostic Extraction Model}

To train the agnostic extraction model, we used a learning rate of 1e-4 with a batch size of 32 and AdamW \cite{loshchilov2019decoupled} optimization algorithm.

\subsubsection{Dense Representations Model}

We use our agnostic extraction model to build the dataset to train and evaluate the model to retrieve 60 candidates for each question-context pair. Each time a good answer was present in the extracted set of candidates, the whole example was added to the train set. To evaluate the model we extract 100 candidates for each question-context pair. \\
The training of the Siamese network took approximately 1 week for 5 epochs on a single 24GB GPU NVIDIA Quadro RTX 6000. \\
We used a learning rate of 1e-5 with AdamW optimizer and a linear scheduler. We also used mixed precision training \cite{micikevicius2018mixed} to reduce time requirements and 8 steps of gradient accumulation along with a batch size of 4, equivalent to a training batch size of 32.

\subsection{Results}

\subsubsection{Answer Candidates Extraction Model}
\begin{figure}[ht]
    \includegraphics[width = 0.4\textwidth]{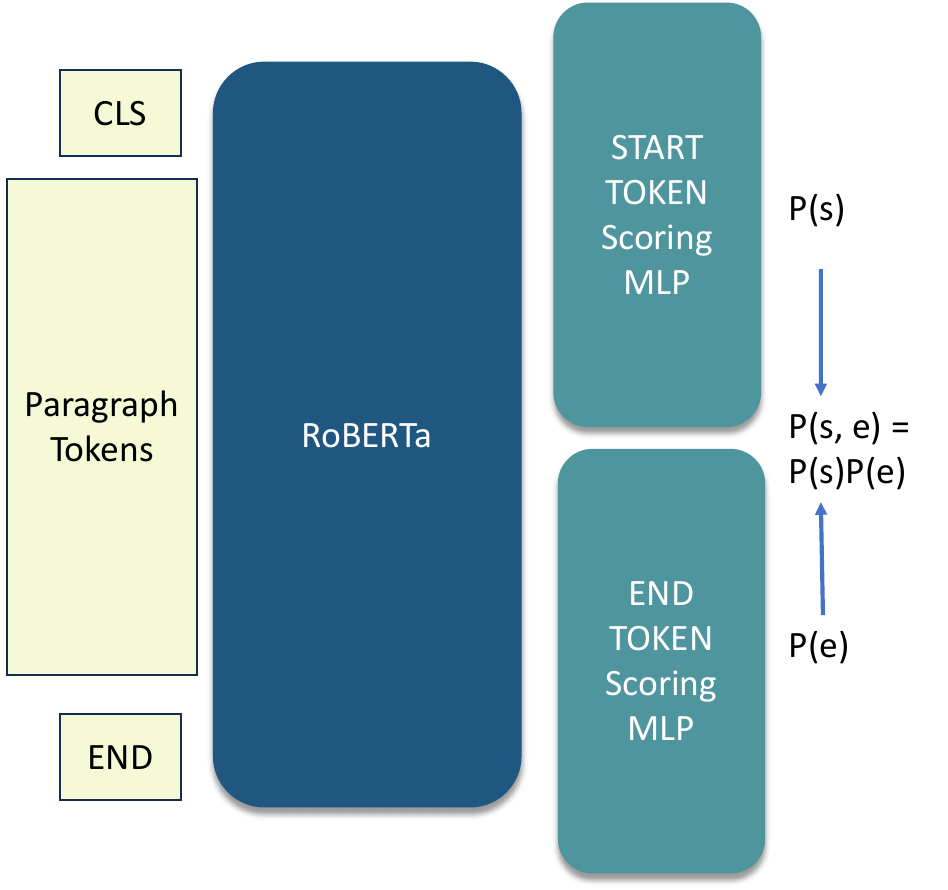}
\caption{classic architecture for candidates extraction}
\label{classic extraction}
\end{figure}

In this section, we justify the architecture of the Answer Candidates Extraction Model. Indeed, we might have chosen a simpler architecture where start positions and end positions likelihoods are computed independently, as shown in figure \ref{classic extraction}. 

While the two models show equivalent results in vanilla Question Answering, our chosen architecture provides a much better set of candidates. Given the same number of selected candidates, the good answer is far more present, as shown in Table 1. The architectures are evaluated with exact-match and f1-score over all selected candidates. We evaluate the classic architecture for fair comparisons using both optimal decoding and beam search. When doing a beam search, we use $50$ as beam size for start positions and $2$ as beam size for end positions. We explain the differences in performances because the dependent computations between start and end positions provide better constituents that are more likely to be answers to questions. Indeed, the likelihood of a candidate is better modeled in this case : 
\[\mathbf{P}(s, e) = \mathbf{P}(s) \times \mathbf{P}(e|s)\] while in the classic architecture : 
\[\mathbf{P}(s, e) = \mathbf{P}(s) \times \mathbf{P}(e)\]

\begin{table}[H]
    \centering
    \begin{tabular}{|c|c|c|}
        \hline
        model & exact-match & f1-score \\
        \hline 
        classic architecture & 65.1 & 80.3\\
        \hline
        classic architecture & 54.1 & 72.4\\
        with beam search & & \\
        \hline
        our architecture & \textbf{92.3} & \textbf{96.7}\\
        \hline
    \end{tabular}
    \caption{Comparison between classic decoding and ours for 100 extracted answer candidates}
    \label{prior results}
\end{table}

\subsubsection{Results on the SQuAD v1.1 benchmark (PIQA challenge)}

Table \ref{squad_results} shows the results obtained by various systems on the PIQA challenge. We observe that EfficientQA beats the previous sota DENSPI (+1.3 in exact-match and + 1.4 in f1-score) while the encoding method of the latter is based on the large version of BERT (340 million parameters), and ours is based on RoBERTa-base (125 million parameters). We can explain these performances by the quality of the representations on the one hand and, on the other hand, by the fact that agnostic extraction drastically reduces the size of the set on which we are looking for the right answer. Hence, it leaves less room for error. 

\begin{table}[ht]
    \centering
      \begin{adjustbox}{max width=0.48\textwidth}
\begin{tabular}{ccc}
    \hline
     & EM & F1 \\
    \hline
    $1^{st}$ baseline : LSTM + SA  & 49.0 & 59.8\\
    \cite{seo2018phraseindexed} & & \\

    $2^{nd}$ baseline : LSTM + SA + ELMO  & 52.7 & 62.7\\
    \cite{seo2018phraseindexed} & & \\
    DENSPI  & \underline{73.6} & \underline{81.7}\\
    \cite{seo-etal-2019-real} & & \\
    Ocean-Q  & 63.0 & 70.5 \\
    \cite{fang2020accelerating} & & \\
    EfficientQA & \textbf{74.9} & \textbf{83.1} \\
    \hline
    RoBERTa (vanilla QA, our run) & 83.0 & 90.4 \\
    \cite{liu2019roberta} & & \\
    \hline
\end{tabular}
\end{adjustbox}
    \caption{Results on SQuAD v1.1}
    \label{squad_results}
\end{table}

\subsubsection{Results on the FQuAD benchmark}

CamemBERT \cite{Martin_2020} is a pretrained French language model based on the RoBERTa architecture. We use it to build the dense representations of the French version of EfficientQA. Table \ref{fquad_results} presents the results of EfficientQA on the FQuAD benchmark. \cite{dhoffschmidt2020fquad} have finetuned CamemBERT to perform vanilla Question Answering on their dataset. The results show that the gap between EfficientQA and finetuned models is closer in English than in French. This might be explained by the volume of data significantly lower in French than in English.

\begin{table}[ht]
    \centering
    \begin{adjustbox}{max width=0.48\textwidth}
    \begin{tabular}{|c|c|c|}
        \hline
        model & exact-match & f1-score \\
        \hline
        EfficientQA (\textbf{PIQA}) & 64.4 & 76.0 \\
        \hline
        \hline
        CamembertQA (\textbf{vanilla QA},  &  &  \\
        our run) & 77.6 & 87.3 \\
         \cite{dhoffschmidt2020fquad} & &\\
    
        \hline
    \end{tabular}
    \end{adjustbox}
    \caption{Performances of EfficientQA on the FQuAD benchmark}
    \label{fquad_results}
\end{table}

\section{Conclusion}
In this paper, we introduced EfficientQA, a phrase-indexed approach to solve question answering. Our system relies on question-agnostic extraction of candidates that reduces the set of possible answers and takes advantage of the pair of sequences type of input of the RoBERTa-base pre-trained language model. EfficientQA achieves state-of-the-art performances on the PIQA benchmark and keeps closing the gap with vanilla Question Answering models, while there is still room for further improvements by using heavier pre-trained language models to build dense representations of questions and candidates. Future research will focus on mobilizing the necessary resources to extend EfficientQA representations to index a whole corpus, such as the entire English Wikipedia, and speed-up open-domain question answering.

\bibliography{main}

\begin{thebibliography}{23}
\expandafter\ifx\csname natexlab\endcsname\relax\def\natexlab#1{#1}\fi

\bibitem[{Bahdanau et~al.(2016)Bahdanau, Cho, and Bengio}]{bahdanau2016neural}
Dzmitry Bahdanau, Kyunghyun Cho, and Yoshua Bengio. 2016.
\newblock \href {http://arxiv.org/abs/1409.0473} {Neural machine translation by jointly learning to align and translate}.

\bibitem[{Chen et~al.(2017)Chen, Fisch, Weston, and Bordes}]{chen2017reading}
Danqi Chen, Adam Fisch, Jason Weston, and Antoine Bordes. 2017.
\newblock \href {http://arxiv.org/abs/1704.00051} {Reading wikipedia to answer open-domain questions}.

\bibitem[{Devlin et~al.(2019)Devlin, Chang, Lee, and Toutanova}]{devlin2019bert}
Jacob Devlin, Ming-Wei Chang, Kenton Lee, and Kristina Toutanova. 2019.
\newblock \href {http://arxiv.org/abs/1810.04805} {Bert: Pre-training of deep bidirectional transformers for language understanding}.

\bibitem[{d'Hoffschmidt et~al.(2020)d'Hoffschmidt, Belblidia, Brendlé, Heinrich, and Vidal}]{dhoffschmidt2020fquad}
Martin d'Hoffschmidt, Wacim Belblidia, Tom Brendlé, Quentin Heinrich, and Maxime Vidal. 2020.
\newblock \href {http://arxiv.org/abs/2002.06071} {Fquad: French question answering dataset}.

\bibitem[{Fang et~al.(2020)Fang, Wang, Gan, Sun, and Liu}]{fang2020accelerating}
Yuwei Fang, Shuohang Wang, Zhe Gan, Siqi Sun, and Jingjing Liu. 2020.
\newblock \href {http://arxiv.org/abs/2009.05167} {Accelerating real-time question answering via question generation}.

\bibitem[{Joshi et~al.(2020)Joshi, Chen, Liu, Weld, Zettlemoyer, and Levy}]{joshi2020spanbert}
Mandar Joshi, Danqi Chen, Yinhan Liu, Daniel~S. Weld, Luke Zettlemoyer, and Omer Levy. 2020.
\newblock \href {http://arxiv.org/abs/1907.10529} {Spanbert: Improving pre-training by representing and predicting spans}.

\bibitem[{Karpukhin et~al.(2020)Karpukhin, Oğuz, Min, Lewis, Wu, Edunov, Chen, and tau Yih}]{karpukhin2020dense}
Vladimir Karpukhin, Barlas Oğuz, Sewon Min, Patrick Lewis, Ledell Wu, Sergey Edunov, Danqi Chen, and Wen tau Yih. 2020.
\newblock \href {http://arxiv.org/abs/2004.04906} {Dense passage retrieval for open-domain question answering}.

\bibitem[{Koch et~al.(2015)Koch, Zemel, and Salakhutdinov}]{Koch2015SiameseNN}
Gregory Koch, Richard Zemel, and Ruslan Salakhutdinov. 2015.
\newblock Siamese neural networks for one-shot image recognition.

\bibitem[{Lee et~al.(2018)Lee, Yun, Kim, Ko, and Kang}]{lee2018ranking}
Jinhyuk Lee, Seongjun Yun, Hyunjae Kim, Miyoung Ko, and Jaewoo Kang. 2018.
\newblock \href {http://arxiv.org/abs/1810.00494} {Ranking paragraphs for improving answer recall in open-domain question answering}.

\bibitem[{Liu et~al.(2019)Liu, Ott, Goyal, Du, Joshi, Chen, Levy, Lewis, Zettlemoyer, and Stoyanov}]{liu2019roberta}
Yinhan Liu, Myle Ott, Naman Goyal, Jingfei Du, Mandar Joshi, Danqi Chen, Omer Levy, Mike Lewis, Luke Zettlemoyer, and Veselin Stoyanov. 2019.
\newblock \href {http://arxiv.org/abs/1907.11692} {Roberta: A robustly optimized bert pretraining approach}.

\bibitem[{Loshchilov and Hutter(2019)}]{loshchilov2019decoupled}
Ilya Loshchilov and Frank Hutter. 2019.
\newblock \href {http://arxiv.org/abs/1711.05101} {Decoupled weight decay regularization}.

\bibitem[{Martin et~al.(2020)Martin, Muller, Ortiz~Suárez, Dupont, Romary, de~la Clergerie, Seddah, and Sagot}]{Martin_2020}
Louis Martin, Benjamin Muller, Pedro~Javier Ortiz~Suárez, Yoann Dupont, Laurent Romary, Éric de~la Clergerie, Djamé Seddah, and Benoît Sagot. 2020.
\newblock \href {https://doi.org/10.18653/v1/2020.acl-main.645} {Camembert: a tasty french language model}.
\newblock \emph{Proceedings of the 58th Annual Meeting of the Association for Computational Linguistics}.

\bibitem[{Micikevicius et~al.(2018)Micikevicius, Narang, Alben, Diamos, Elsen, Garcia, Ginsburg, Houston, Kuchaiev, Venkatesh, and Wu}]{micikevicius2018mixed}
Paulius Micikevicius, Sharan Narang, Jonah Alben, Gregory Diamos, Erich Elsen, David Garcia, Boris Ginsburg, Michael Houston, Oleksii Kuchaiev, Ganesh Venkatesh, and Hao Wu. 2018.
\newblock \href {http://arxiv.org/abs/1710.03740} {Mixed precision training}.

\bibitem[{Min et~al.(2018)Min, Zhong, Socher, and Xiong}]{min2018efficient}
Sewon Min, Victor Zhong, Richard Socher, and Caiming Xiong. 2018.
\newblock \href {http://arxiv.org/abs/1805.08092} {Efficient and robust question answering from minimal context over documents}.

\bibitem[{Raison et~al.(2018)Raison, Mazaré, Das, and Bordes}]{raison2018weaver}
Martin Raison, Pierre-Emmanuel Mazaré, Rajarshi Das, and Antoine Bordes. 2018.
\newblock \href {http://arxiv.org/abs/1804.10490} {Weaver: Deep co-encoding of questions and documents for machine reading}.

\bibitem[{Rajpurkar et~al.(2016)Rajpurkar, Zhang, Lopyrev, and Liang}]{rajpurkar2016squad}
Pranav Rajpurkar, Jian Zhang, Konstantin Lopyrev, and Percy Liang. 2016.
\newblock \href {http://arxiv.org/abs/1606.05250} {Squad: 100,000+ questions for machine comprehension of text}.

\bibitem[{Robertson and Jones(1976)}]{https://doi.org/10.1002/asi.4630270302}
S.~E. Robertson and K.~Sparck Jones. 1976.
\newblock \href {https://doi.org/https://doi.org/10.1002/asi.4630270302} {Relevance weighting of search terms}.
\newblock \emph{Journal of the American Society for Information Science}, 27(3):129--146.

\bibitem[{Seo et~al.(2018{\natexlab{a}})Seo, Kembhavi, Farhadi, and Hajishirzi}]{seo2018bidirectional}
Minjoon Seo, Aniruddha Kembhavi, Ali Farhadi, and Hannaneh Hajishirzi. 2018{\natexlab{a}}.
\newblock \href {http://arxiv.org/abs/1611.01603} {Bidirectional attention flow for machine comprehension}.

\bibitem[{Seo et~al.(2018{\natexlab{b}})Seo, Kwiatkowski, Parikh, Farhadi, and Hajishirzi}]{seo2018phraseindexed}
Minjoon Seo, Tom Kwiatkowski, Ankur~P. Parikh, Ali Farhadi, and Hannaneh Hajishirzi. 2018{\natexlab{b}}.
\newblock \href {http://arxiv.org/abs/1804.07726} {Phrase-indexed question answering: A new challenge for scalable document comprehension}.

\bibitem[{Seo et~al.(2019)Seo, Lee, Kwiatkowski, Parikh, Farhadi, and Hajishirzi}]{seo-etal-2019-real}
Minjoon Seo, Jinhyuk Lee, Tom Kwiatkowski, Ankur Parikh, Ali Farhadi, and Hannaneh Hajishirzi. 2019.
\newblock \href {https://doi.org/10.18653/v1/P19-1436} {Real-time open-domain question answering with dense-sparse phrase index}.
\newblock In \emph{Proceedings of the 57th Annual Meeting of the Association for Computational Linguistics}, pages 4430--4441, Florence, Italy. Association for Computational Linguistics.

\bibitem[{Sparck~Jones(1988)}]{10.5555/106765.106782}
Karen Sparck~Jones. 1988.
\newblock \emph{A Statistical Interpretation of Term Specificity and Its Application in Retrieval}, page 132–142. Taylor Graham Publishing, GBR.

\bibitem[{Wang et~al.(2017)Wang, Yu, Guo, Wang, Klinger, Zhang, Chang, Tesauro, Zhou, and Jiang}]{wang2017r3}
Shuohang Wang, Mo~Yu, Xiaoxiao Guo, Zhiguo Wang, Tim Klinger, Wei Zhang, Shiyu Chang, Gerald Tesauro, Bowen Zhou, and Jing Jiang. 2017.
\newblock \href {http://arxiv.org/abs/1709.00023} {R$^3$: Reinforced reader-ranker for open-domain question answering}.

\bibitem[{Yang et~al.(2019)Yang, Xie, Lin, Li, Tan, Xiong, Li, and Lin}]{Yang_2019}
Wei Yang, Yuqing Xie, Aileen Lin, Xingyu Li, Luchen Tan, Kun Xiong, Ming Li, and Jimmy Lin. 2019.
\newblock \href {https://doi.org/10.18653/v1/n19-4013} {End-to-end open-domain question answering with}.
\newblock \emph{Proceedings of the 2019 Conference of the North}.

\end{thebibliography}
\bibliographystyle{acl_natbib}

\end{document}